# NuMVC: An Efficient Local Search Algorithm
# for Minimum Vertex Cover

**Shaowei Cai**                                          SHAOWEICAI.CS@GMAIL.COM
*Key Laboratory of High Confidence Software Technologies*
*Peking University, Beijing, China*

**Kaile Su**                                             K.SU@GRIFFITH.EDU.AU
*Institute for Integrated and Intelligent Systems*
*Griffith University, Brisbane, Australia*

**Chuan Luo**                                            CHUANLUOSABER@GMAIL.COM
*Key Laboratory of High Confidence Software Technologies*
*Peking University, Beijing, China*

**Abdul Sattar**                                         A.SATTAR@GRIFFITH.EDU.AU
*Institute for Integrated and Intelligent Systems*
*Griffith University, Brisbane, Australia*

## Abstract

The Minimum Vertex Cover (MVC) problem is a prominent NP-hard combinatorial optimization problem of great importance in both theory and application. Local search has proved successful for this problem. However, there are two main drawbacks in state-of-the-art MVC local search algorithms. First, they select a pair of vertices to exchange simultaneously, which is time-consuming. Secondly, although using edge weighting techniques to diversify the search, these algorithms lack mechanisms for decreasing the weights. To address these issues, we propose two new strategies: two-stage exchange and edge weighting with forgetting. The two-stage exchange strategy selects two vertices to exchange separately and performs the exchange in two stages. The strategy of edge weighting with forgetting not only increases weights of uncovered edges, but also decreases some weights for each edge periodically. These two strategies are used in designing a new MVC local search algorithm, which is referred to as NuMVC.

We conduct extensive experimental studies on the standard benchmarks, namely DIMACS and BHOSLIB. The experiment comparing NuMVC with state-of-the-art heuristic algorithms show that NuMVC is at least competitive with the nearest competitor namely PLS on the DIMACS benchmark, and clearly dominates all competitors on the BHOSLIB benchmark. Also, experimental results indicate that NuMVC finds an optimal solution much faster than the current best exact algorithm for Maximum Clique on random instances as well as some structured ones. Moreover, we study the effectiveness of the two strategies and the run-time behaviour through experimental analysis.

## 1. Introduction

The Minimum Vertex Cover (MVC) problem consists of, given an undirected graph $G = (V, E)$, finding the minimum sized vertex cover, where a vertex cover is a subset $S \subseteq V$ such that every edge in $G$ has at least one endpoint in $S$. MVC is an important combinatorial optimization problem with many real-world applications, such as network security, scheduling, VLSI design and industrial machine assignment. It is equivalent to two other well-known combinatorial optimization problems: the Maximum Independent Set (MIS) problem and the Maximum Clique (MC) problem, which have a wide range of applications in areas such as information retrieval, experimental design, signal transmission, computer vision, and also bioinformatics problems such as aligning DNA and protein sequences (Johnson & Trick, 1996). Indeed, these three problems can be seen as three different forms of the same problem, from the viewpoint of practical algorithms. Algorithms for MVC can be directly used to solve the MIS and MC problems. Due to their great importance in theory and applications, these three problems have been widely investigated for the last several decades (Carraghan & Pardalos, 1990; Evans, 1998; Pullan & Hoos, 2006; Richter, Helmert, & Gretton, 2007; Cai, Su, & Chen, 2010; Li & Quan, 2010b; Cai, Su, & Sattar, 2011).

Theoretical analyses indicate that these three problems MVC, MIS, and MC are computationally hard. They are all NP-hard and the associated decision problems are NP-complete (Garey & Johnson, 1979). Moreover, they are hard to solve approximately. It is NP-hard to approximate MVC within any factor smaller than 1.3606 (Dinur & Safra, 2005), although one can achieve an approximation ratio of $2 - o(1)$ (Halperin, 2002; Karakostas, 2005). Besides the inapproximability of MVC, Håstad shows that both MIS and MC are not approximable within $|V|^{1-\epsilon}$ for any $\epsilon > 0$, unless NP=ZPP[1] (Håstad, 1999, 2001). Recently, this conclusion has been enhanced that MC is not approximable within $|V|^{1-\epsilon}$ for any $\epsilon > 0$ unless NP=P (Zuckerman, 2006), derived from a derandomization of Håstad's result. Moreover, the currently best polynomial-time approximation algorithm for MC is only guaranteed to find a clique within a factor of $O(n(log log n)^2/(log n)^3)$ of optimum (Feige, 2004).

The algorithms to solve MVC (MIS, MC) fall into two types: exact algorithms and heuristic algorithms. Exact methods which mainly include branch-and-bound algorithms (Carraghan & Pardalos, 1990; Fahle, 2002; Östergård, 2002; Régin, 2003; Tomita & Kameda, 2009; Li & Quan, 2010b, 2010a), guarantee the optimality of the solutions they find, but may fail to give a solution within reasonable time for large instances. Heuristic algorithms, which mainly include local search algorithms, cannot guarantee the optimality of their solutions, but they can find optimal or satisfactory near-optimal solutions for large and hard instances within reasonable time. Therefore, it is appealing to use local search algorithms to solve large and hard MVC (MC, MIS) instances.

Early heuristic methods for Maximum Clique have been designed as initial responses to the Second DIMACS Implementation Challenge (Johnson & Trick, 1996), where Maximum Clique is one of the three challenge problems. After that, a huge amount of effort was devoted to designing local search algorithms for MVC, MC and MIS problems (Aggarwal, Orlin, & Tai, 1997; Battiti & Protasi, 2001; Busygin, Butenko, & Pardalos, 2002; Shyu, Yin, & Lin, 2004; Barbosa & Campos, 2004; Pullan, 2006; Richter et al., 2007; Andrade, Resende, & Werneck, 2008; Cai et al., 2010, 2011). A review of heuristic algorithms for these three problems can be found in a recent paper on MVC local search (Cai et al., 2011).

This work is devoted to a more efficient local search algorithm for MVC. Typically, local search algorithms for MVC solve the problem by iteratively solving the $k$-vertex cover problem. To solve the $k$-vertex cover problem, they maintain a current candidate solution of size $k$, and exchange two vertices iteratively until it becomes a vertex cover. However, we observe two drawbacks in state-of-the-art MVC local search algorithms. First, they select a pair of vertices for exchanging simultaneously according to some heuristic (Richter et al., 2007; Cai et al., 2010, 2011), which is rather time-consuming, as will be explained in Section 3. The second drawback is about the edge weighting techniques. The basic concept of edge weighting is to increase weights of uncovered edges to diversify the search. Previous MVC local search algorithms utilize different edge weighting schemes. For example, COVER (Richter et al., 2007) increases weights of uncovered edges at each step, while EWLS (Cai et al., 2010) and EWCC (Cai et al., 2011) increase weights of uncovered edges only when reaching local optima. However, all these algorithms do not have a mechanism to decrease the weights. We believe this is deficient because the weighting decisions made too long ago may mislead the search.

---

1. ZPP is the class of problems that can be solved in expected polynomial time by a probabilistic algorithm with zero error probability.

To address these two issues in MVC local search algorithms, this paper proposes two new strategies, namely two-stage exchange and edge weighting with forgetting. The two-stage exchange strategy decomposes the exchanging procedure into two stages, i.e., the removing stage and the adding stage, and performs them separately. It first selects a vertex and removes it from the current candidate solution, and then selects a vertex in a random uncovered edge and adds it. The two-stage exchange strategy yields an efficient two-pass move operator for MVC local search, in which the first pass is a linear-time search for the vertex-to-remove, while the second pass is a linear-time search for the vertex-to-add. This is in contrast to the standard quadratic, all-at-once move operator. Moreover, the two-stage exchange strategy renders the algorithm more flexible in that we can employ different heuristics in different stages. Indeed, the NuMVC algorithm utilizes a highly greedy heuristic for the removing stage, while for the adding stage, it makes good use of a diversifying heuristic within a framework similar to focused random walk (Papadimitriou, 1991).

The second strategy we propose is edge weighting with forgetting. It increases weights of uncovered edges by one at each step. Moreover, when the averaged edge weight achieves a threshold, it reduces weights of all edges by multiplying a constant factor $\rho$ ($0 < \rho < 1$) to forget the earlier weighting decisions. To the best of our knowledge, this is the first time a forgetting mechanism is introduced into local search algorithms for MVC.

The two strategies are combined to design a new local search algorithm called NuMVC. We carry out a detailed experimental study to investigate the performance of NuMVC, and compare it with PLS (Pullan, 2006), COVER (Richter et al., 2007) and EWCC (Cai et al., 2011), which are leading heuristic algorithms for MVC (MC, MIS). Experimental results show that NuMVC competes well with other solvers on the DIMACS benchmark, and shows a dramatic improvement over existing results on the whole BHOSLIB benchmark. These parts of work have been published in an early version of this paper (Cai, Su, & Sattar, 2012).

In this paper, we additionally carry out more experimental analyses and provides further insights about the two strategies in NuMVC. We compare NuMVC with the exact algorithm MaxCLQdyn+EFL+SCR (Li & Quan, 2010a), which is the best exact Maximum Clique algorithm we found in the literature. Experimental results indicate that NuMVC finds an optimal solution much faster than the exact algorithm on random instances as well as some structured ones. More importantly, we conduct experimental investigations to study the run-time behaviour of NuMVC and the effectiveness of the two new strategies in NuMVC.

The remainder of this paper is organized as follows. In the next section, we introduce some definitions and notations used in this paper. We then present the two strategies: two-stage exchange and edge weighting with forgetting. In Section 5, we describe the NuMVC algorithm. Section 6 presents the experimental study of NuMVC and comparative results to other algorithms, including heuristic and exact algorithms. This is followed by more detailed investigations about the run-time behaviour of NuMVC and the effectiveness of the two new strategies in Section 7. Finally, we conclude the paper by summarizing the main contributions and some future directions.

## 2. Preliminaries

An undirected graph $G = (V, E)$ consists of a *vertex set $V$* and an *edge set $E \subseteq V \times V$*, where each edge is a 2-element subset of $V$. For an edge $e = \{u, v\}$, we say that vertices $u$ and $v$ are the *endpoints* of edge $e$. Two vertices are neighbors if and only if they both belong to some common edge. We denote $N(v) = \{u \in V | \{u, v\} \in E\}$, the set of neighbors of a vertex $v$.

For an undirected graph $G = (V, E)$, an *independent set* is a subset of $V$ with pairwise non-adjacent elements and a *clique* is a subset of $V$ with pairwise adjacent elements. The *maximum independent set* and *maximum clique* problems are to find the maximum sized independent set and clique in a graph, respectively.

We note that these three problems MVC, MIS and MC can be seen as three different forms of the same problem, from the viewpoint of experimental algorithms. A vertex set $S$ is an independent set of $G$ if and only if $V \setminus S$ is a vertex cover of $G$; a vertex set $K$ is a clique of $G$ if and only if $V \setminus K$ is a vertex cover of the complementary graph $\overline{G}$. To find the maximum independent set of a graph $G$, one can find the minimum vertex cover $C_{min}$ for $G$ and return $V \setminus C_{min}$. Similarly, to find the maximum clique of a graph $G$, one can find the minimum vertex cover $C'_{min}$ for the complementary graph $\overline{G}$, and return $V \setminus C'_{min}$.

Given an undirected graph $G = (V, E)$, a *candidate solution* for MVC is a subset of vertices. An edge $e \in E$ is *covered* by a candidate solution $X$ if at least one endpoint of $e$ belongs to $X$. During the search procedure, NuMVC always maintains a current candidate solution. For convenience, in the rest of this paper, we use $C$ to denote the current candidate solution. The *state* of a vertex $v$ is denoted by $s_v \in \{1, 0\}$, such that $s_v = 1$ means $v \in C$, and $s_v = 0$ means $v \notin C$. The step to a neighboring candidate solution consists of exchanging two vertices: a vertex $u \in C$ is removed from $C$, and a vertex $v \notin C$ is put into $C$. The *age* of a vertex is the number of steps since its state was last changed.

As with most state-of-the-art MVC local search algorithms, NuMVC utilizes an edge weighting scheme. For edge weighting local search, we follow the definitions and notations in EWCC (Cai et al., 2011). An edge weighted undirected graph is an undirected graph $G = (V, E)$ combined with a weighting function $w$ so that each edge $e \in E$ is associated with a non-negative integer number $w(e)$ as its weight. We use $\overline{w}$ to denote the mean value of all edge weights.

Let $w$ be a weighting function for $G$. For a candidate solution $X$, we set the cost of $X$ as

$$cost(G, X) = \sum_{e \in E \text{ and } e \text{ is not covered by } X} w(e)$$

which indicates the total weight of edges uncovered by $X$. We take $cost(G, X)$ as the *evaluation function*, and NuMVC prefers candidate solutions with lower costs.

For a vertex $v \in V$,

$$dscore(v) = cost(G, C) - cost(G, C')$$

where $C' = C \setminus \{v\}$ if $v \in C$, and $C' = C \cup \{v\}$ otherwise, measuring the benefit of changing the state of vertex $v$. Obviously, for a vertex $v \in C$, we have $dscore(v) \leq 0$, and the greater *dscore* indicates the less loss of covered edges by removing it out of $C$. For a vertex $v \notin C$, we have $dscore(v) \geq 0$, and the higher *dscore* indicates the greater increment of covered edges by adding it into $C$.

## 3. Two-Stage Exchange

In this section, we introduce the two-stage exchange strategy, which is adopted by the NuMVC algorithm to exchange a pair of vertices.

As with most state-of-the-art MVC local search algorithms, NuMVC is an iterated $k$-vertex cover algorithm. When finding a $k$-vertex cover, NuMVC removes one vertex from the current candidate solution $C$ and goes on to search for a $(k - 1)$-vertex cover. In this sense, the core of NuMVC is a $k$-vertex cover algorithm — given a positive integer number $k$, searching for a $k$-sized vertex cover. To find a $k$-vertex cover, NuMVC begins with a candidate solution $C$ of size $k$, and exchanges two vertices iteratively until $C$ becomes a vertex cover.

Most local search algorithms for MVC select a pair of vertices to exchange simultaneously according to a certain heuristic. For example, COVER selects a pair of vertices that maximize $gain(u, v)$ (Richter et al., 2007), while EWLS (Cai et al., 2010) and EWCC (Cai et al., 2011) select a random pair of vertices with $score(u, v) > 0$. This strategy of selecting two vertices to exchange simultaneously leads to a quadratic neighborhood for candidate solutions. Moreover, the evaluation of a pair of vertices not only depends on the evaluations (such as $dscore$) of the two vertices, but also involves the relationship between the two vertices, like "do they belong to a same edge". Therefore, it is rather time-consuming to evaluate all candidate pairs of vertices.

In contrast to earlier MVC local search algorithms, NuMVC selects the two vertices for exchanging separately and exchanges the two selected vertices in two stages. In each iteration, NuMVC first selects a vertex $u \in C$ with the highest $dscore$ and removes it. After that, NuMVC selects a uniformly random uncovered edge $e$, and chooses one endpoint $v$ of $e$ with the higher $dscore$ under some restrictions and adds it into $C$. Note that this two-stage exchange strategy resembles in some respect the min-conflicts hill-climbing heuristic for CSP (Minton, Johnston, Philips, & Laird, 1992), which shows surprisingly good performance for the N-queens problem.

Selecting the two vertices for exchanging separately may in some cases miss some greedier vertex pairs which consist of two neighboring vertices. However, as is usual in local search algorithms, there is a trade-off between the accuracy of heuristics and the complexity per step. Let $R$ and $A$ denote the set of candidate vertices for removing and adding separately. The time complexity per step for selecting the exchanging vertex pair simultaneously is $|R| \cdot |A|$; while the complexity per step for selecting the two vertices separately, as in NuMVC, is only $|R| + |A|$. It is worthy to note that, as heuristics in a local search algorithm are often based on intuition and experience rather than on theoretically or empirically derived principles and insights, we cannot say for certain that being less greedy is not a good thing (Hoos & Stützle, 2004). On the other hand, a lower time complexity is always desirable.

## 4. Edge Weighting with Forgetting

In this section, we present a new edge weighting technique called edge weighting with forgetting, which plays an important role in NuMVC.

The proposed strategy of edge weighting with forgetting works as follows. Each edge is associated with a positive integer number as its weight, and each edge weight is initialized as one. Then in each iteration, edge weights of the uncovered edges are increased by one. Moreover, when the average weight achieves a threshold, all edge weights are reduced to forget the earlier weighting decisions using the formula $w(e) := \lfloor \rho \cdot w(e) \rfloor$, where $\rho$ is a constant factor between 0 and 1.

Note that edge weighting techniques in MVC local search, including the one in this work, fall in the more general penalty idea for optimization problems, which dates back to Morris 's breakout method (Morris, 1993) and has been widely used in local search algorithms for constraint optimization problems such as SAT (Yugami, Ohta, & Hara, 1994; Wu & Wah, 2000; Schuurmans, Southey, & Holte, 2001; Hutter, Tompkins, & Hoos, 2002). Our results therefore provide further evidence for the effectiveness and general applicability of this algorithmic technique.

Edge weighting techniques have been successfully used to improve MVC local search algorithms. For example, COVER (Richter et al., 2007) updates edge weights at each step, while EWLS (Cai et al., 2010) and EWCC (Cai et al., 2011) update edge weights only when reaching local optima. However, all previous edge weighting techniques do not have a mechanism to decrease the weights, which limits their effectiveness. The strategy of edge weighting with forgetting in this work introduces a forgetting mechanism to reduce edge weights periodically, which contributes considerably to the NuMVC algorithm.

The intuition behind the forgetting mechanism is that the weighting decisions made too long ago are no longer helpful and may mislead the search, and hence should be considered less important than the recent ones. For example, consider two edges $e_1$ and $e_2$ with $w(e_1) = 1000$ and $w(e_2) = 100$ at some step. We use $\Delta w(e)$ to denote the increase of $w(e)$. According to the *evaluation function*, in the next period of time, the algorithm is likely to cover $e_1$ more frequently than $e_2$, and we may assume that during this period $\Delta w(e_1) = 50$ and $\Delta w(e_2) = 500$, which makes $w(e_1) = 1000 + 50 = 1050$ and $w(e_2) = 100 + 500 = 600$. Without a forgetting mechanism, the algorithm would still prefer $e_1$ to $e_2$ to be covered in the future search. This is not reasonable, as during this period $e_2$ is covered in much fewer steps than $e_1$ is. Thus, $e_2$ should take priority to be covered for the sake of diversification. Now let us consider the case with the forgetting mechanism (assuming $\rho = 0.3$ which is the setting in our experiments). Suppose $w(e_1) = 1000$ and $w(e_2) = 100$ when the algorithm performs the forgetting. The forgetting mechanism reduces the edge weights as $w(e_1) = 1000 \times 0.3 = 300$ and $w(e_2) = 100 \times 0.3 = 30$. After a period of time, with $\Delta w(e_1) = 50$ and $\Delta w(e_2) = 500$, we have $w(e_1) = 300 + 50 = 350$ and $w(e_2) = 30 + 500 = 530$. In this case, the algorithm prefers to cover $e_2$ rather than cover $e_1$ in the future search, as we expect.

Although being inspired by smoothing techniques in clause weighting local search algorithms for SAT, the forgetting mechanism in NuMVC differs from those smoothing techniques in SAT local search algorithms. According to the way that clause weights are smoothed, there are three main smoothing techniques in clause weighting local search algorithms for SAT to the best of our knowledge: the first is to pull all clause weights to their mean value using the formula $w_i := \rho \cdot w_i + (1 - \rho) \cdot \overline{w}$, as in ESG (Schuurmans et al., 2001), SAPS (Hutter et al., 2002) and Swcca (Cai & Su, 2012); the second is to subtract one from all clause weights which are greater than one, as in DLM (Wu & Wah, 2000) and PAWS (Thornton, Pham, Bain, & Jr., 2004); and the last is employed in DDWF (Ishtaiwi, Thornton, Sattar, & Pham, 2005), which transfers weights from neighbouring satisfied clauses to unsatisfied ones. It is obvious that the forgetting mechanism in NuMVC is different from all these smoothing techniques.

Recently, a forgetting mechanism was proposed for vertex weighting technique in the significant MC local search algorithm DLS-MC (Pullan & Hoos, 2006), which is an important sub-algorithm in PLS (Pullan, 2006) and CLS (Pullan, Mascia, & Brunato, 2011). The DLS-MC algorithm employs a vertex weighting scheme which increases the weights of vertices (by one) not in the current clique when reaching a local optimum, and periodically decreases weights (by one) for all vertices that currently have a penalty. Specifically, it utilizes a parameter $pd$ (*penalty delay*) to specify the number of penalty increase iterations that must occur before the algorithm performs a forgetting operation. However, Pullan and Hoos also observed that DLS-MC is very sensitive to the $pd$ parameter, and the optimal value of $pd$ varies considerably among different instances. Indeed, the performance of DLS-MC in is given by optimizing the $pd$ parameter. In contrast, the forgetting mechanism in NuMVC is much less sensitive to its parameters (as will be shown in Section 7.4), and thus is more robust.

We also notice that the formula used in the forgetting mechanism in NuMVC has been adopted in long-term frequency-based learning mechanisms for tabu search (Taillard, 1994). However, in Taillar's algorithm, the parameter $\rho$ (using the term in this work) is always greater than one, and the formula is used for penalizing a move rather than forgetting the penalties.

## 5. The NuMVC Algorithm

In this section, we present the NuMVC algorithm, which utilizes the strategies of two-stage exchange and edge weighting with forgetting.

---

**Algorithm 1**: NuMVC

1   NuMVC ($G$,*cutoff*)
    **Input**: graph $G = (V, E)$, the *cutoff* time
    **Output**: vertex cover of $G$
2   **begin**
3      initialize edge weights and $dscores$ of vertices;
4      initialize the *confChange* array as an all-1 array;
5      construct $C$ greedily until it is a vertex cover;
6      $C^* := C$;
7      **while** *elapsed time < cutoff* **do**
8          **if** *there is no uncovered edge* **then**
9              $C^* := C$;
10              remove a vertex with the highest $dscore$ from $C$;
11              continue;
12          choose a vertex $u \in C$ with the highest $dscore$, breaking ties in favor of the oldest one;
13          $C := C \backslash \{u\}$, *confChange*$(u) := 0$ and *confChange*$(z) := 1$ for each $z \in N(u)$;
14          choose an uncovered edge $e$ randomly;
15          choose a vertex $v \in e$ such that *confChange*$(v) = 1$ with higher $dscore$, breaking ties in favor of the older one;
16          $C := C \cup \{v\}$, *confChange*$(z) := 1$ for each $z \in N(v)$;
17          $w(e) := w(e) + 1$ for each uncovered edge $e$;
18          **if** $\overline{w} \geq \gamma$ **then** $w(e) := \lfloor \rho \cdot w(e) \rfloor$ for each edge $e$;
19      **return** $C^*$;
20 **end**

---

For better understanding the algorithm, we first describe a strategy called configuration checking (CC), which is used in NuMVC. The CC strategy (Cai et al., 2011) was proposed for handling the cycling problem in local search, i.e., revisiting a candidate solution that has been visited recently (Michiels, Aarts, & Korst, 2007). This strategy has been successfully applied in local search algorithms for MVC (Cai et al., 2011) as well as SAT (Cai & Su, 2011, 2012).

The CC strategy in NuMVC works as follows: For a vertex $v \notin C$, if all its neighboring vertices never change their *states* since the last time $v$ was removed from $C$, then $v$ should not be added back to $C$. The CC strategy can be seen as a prohibition mechanism, which shares the same spirit but differs from the well-known prohibition mechanism called tabu (Glover, 1989).

An implementation of the CC strategy is to maintain a Boolean array *confChange* for vertices. During the search procedure, those vertices which have a *confChange* value of 0 are forbidden to add into $C$. The *confChange* array is initialized as an all-1 array. After that, when a vertex $v$ is removed from $C$, $confChange(v)$ is reset to 0, and when a vertex $v$ changes its *state*, for each $z \in N(v)$, $confChange(z)$ is set to 1.

We outline the NuMVC algorithm in Algorithm 1, as described below. In the beginning, all edge weights are initialized as 1, and $dscores$ of vertices are computed accordingly; $confChange(v)$ is initialized as 1 for each vertex $v$; then the current candidate solution $C$ is constructed by iteratively adding the vertex with the highest $dscore$ (ties are broken randomly), until it becomes a vertex cover. Finally, the best solution $C^*$ is initialized as $C$.

After the initialization, the loop (lines 7-18) is executed until a given cutoff time is reached. During the search procedure, once there is no uncovered edge, which means $C$ is a vertex cover, NuMVC updates the best solution $C^*$ as $C$ (line 9). Then it removes one vertex with the highest $dscore$ from $C$ (line 10), breaking ties randomly, so that it can go on to search for a vertex cover of size $|C| = |C^*| - 1$. We note that, in $C$, the vertex with the highest $dscore$ has the minimum absolute value of $dscore$ since all these $dscores$ are negative.

In each iteration of the loop, NuMVC swaps two vertices according to the strategy of two-stage exchange (lines 12-16). Specifically, it first selects a vertex $u \in C$ with the highest $dscore$ to remove, breaking ties in favor of the oldest one. After removing $u$, NuMVC chooses an uncovered edge $e$ uniformly at random, and selects one of $e$'s endpoints to add into $C$ as follows: If there is only one endpoint whose *confChange* is 1, then that vertex is selected; if the *confChange* values of both endpoints are 1, then NuMVC selects the vertex with the higher $dscore$, breaking ties in favor of the older one. The exchange is finished by adding the selected vertex into $C$. Along with exchanging the two selected vertices, the *confChange* array is updated accordingly.

At the end of each iteration, NuMVC updates the edge weights (lines 17-18). First, weights of all uncovered edges are increased by one. Moreover, NuMVC utilizes the forgetting mechanism to decrease the weights periodically. In detail, if the averaged weight of all edges achieves a threshold $\gamma$, then all edge weights are multiplied by a constant factor $\rho$ ($0 < \rho < 1$) and rounded down to an integer as edge weights are defined as integers in NuMVC. The forgetting mechanism forgets the earlier weighting decision to some extent, as these past effects are generally no longer helpful and may mislead the search.

We conclude this section by the following observation, which guarantees the executability of line 15.

**Proposition 1.** *For an uncovered edge $e$, there is at least one endpoint $v$ of edge $e$ such that confChange$(v) = 1$.*

**Proof:** Let us consider an arbitrary uncovered edge $e = \{v_1, v_2\}$. The proof includes two cases.
(a) *There is at least one of $v_1$ and $v_2$ which never changes its state after initialization.* Without

loss of generality, we assume $v_1$ is such a vertex. In the initialization, *confChange*($v_1$) is set to 1. After that, only removing $v_1$ from $C$ (which corresponds to $v$'s state $s_v$ changing to 0 from 1) can make *confChange*($v_1$) be 0, but $v_1$ never changes its state after initialization, so we have *confChange*($v_1$)= 1.

(b) *Both $v_1$ and $v_2$ change their states after initialization.* As $e$ is uncovered, we have $v_1 \notin C$ and $v_2 \notin C$. Without loss of generality, we assume the last removing of $v_1$ happens before the last removing of $v_2$. The last time $v_1$ is removed, $v_2 \in C$ holds. Afterwards, $v_2$ is removed, which means $v_2$ changes its state, so *confChange*($v_1$) is set to 1 as $v_1 \in N(v_2)$. □

## 6. Empirical Results

In this section, we present a detailed experimental study to evaluate the performance of NuMVC on standard benchmarks in the literature, i.e., the DIMACS and BHOSLIB benchmarks. We first introduce the DIMACS and BHOSLIB benchmarks, and describe some preliminaries about the experiments. Then, we divide the experiments into three parts. The purpose of the first part is to demonstrate the performance of NuMVC in detail. The second is to compare NuMVC with state-of-the-art heuristic algorithms. Finally, the last part is to compare NuMVC with state-of-the-art exact algorithms.

### 6.1 The Benchmarks

Having a good set of benchmarks is fundamental to demonstrate the effectiveness of new solvers. We use the two standard benchmarks in MVC (MIS, MC) research, the DIMACS benchmark and the BHOSLIB benchmark. The DIMACS benchmark includes instances from industry and those generated by various models, while the BHOSLIB instances are random ones of high difficulty.

#### 6.1.1 DIMACS Benchmark

The DIMACS benchmark is taken from the Second DIMACS Implementation Challenge for the Maximum Clique problem (1992-1993)[2]. Thirty seven graphs were selected by the organizers for a summary to indicate the effectiveness of algorithms, comprising the Second DIMACS Challenge Test Problems. These instances were generated from real world problems such as coding theory, fault diagnosis, Keller's conjecture and the Steiner Triple Problem, etc, and random graphs in various models, such as the `brock` and `p_hat` families. These instances range in size from less than 50 vertices and 1,000 edges to greater than 4,000 vertices and 5,000,000 edges. Although being proposed two decades ago, the DIMACS benchmark remains the most popular benchmark and has been widely used for evaluating heuristic algorithms for MVC (Richter et al., 2007; Pullan, 2009; Cai et al., 2011; Gajurel & Bielefeld, 2012), MIS (Andrade et al., 2008; Pullan, 2009) and MC algorithms (Pullan, 2006; Katayama, Sadamatsu, & Narihisa, 2007; Grosso, Locatelli, & Pullan, 2008; Pullan et al., 2011; Wu, Hao, & Glover, 2012). In particular, the DIMACS benchmark has been used for evaluating COVER and EWCC. It is convenient for us to use this benchmark also to conduct experiments comparing NuMVC with COVER and EWCC. Note that as the DIMACS graphs were originally designed for the Maximum Clique problem, MVC algorithms are tested on their complementary graphs.

---

### 6.1.2 BHOSLIB BENCHMARK

The BHOSLIB[3] (Benchmarks with Hidden Optimum Solutions) instances were generated randomly in the phase transition area according to the model RB (Xu, Boussemart, Hemery, & Lecoutre, 2005). Generally, those phase-transition instances generated by model RB have been proved to be hard both theoretically (Xu & Li, 2006) and practically (Xu & Li, 2000; Xu, Boussemart, Hemery, & Lecoutre, 2007). The SAT version of the BHOSLIB benchmark is extensively used in the SAT competitions[4]. Nevertheless, SAT solvers are much weaker than MVC solvers on these problems, which remains justifiable when referring to the results of SAT Competition 2011 on this benchmark. The BHOSLIB benchmark is famous for its hardness and influential enough to be strongly recommended by the MVC (MC, MIS) community (Grosso et al., 2008; Cai et al., 2011). It has been widely used in the recent literature as a reference point for new local search solvers to MVC, MC and MIS[5]. Besides these 40 instances, there is a large instance `frb100-40` with 4,000 vertices and 572,774 edges, which is designed for challenging MVC (MC, MIS) algorithms.

The BHOSLIB benchmark was designed for MC, MVC and MIS, and all the graphs in this benchmark are expressed in two formats, i.e., the `clq` format and the `mis` format. For a BHOSLIB instance, the graph in `clq` format and the one in `mis` format are complementary to each other. MC algorithms are tested on the graphs in `clq` format, while MVC and MIS algorithms are tested on those in `mis` format.

## 6.2 Experiment Preliminaries

Before we discuss the experimental results, let us introduce some preliminary information about our experiments.

NuMVC is implemented in C++. The codes of both NuMVC and EWCC are publicly available on the first author's homepage[6]. The codes of COVER are downloaded online[7], and those of PLS are kindly provided by its authors. All the four solvers are compiled by g++ with the '-O2' option. All experiments are carried out on a machine with a 3 GHz Intel Core 2 Duo CPU E8400 and 4GB RAM under Linux. To execute the DIMACS machine benchmarks[8], this machine requires 0.19 CPU seconds for r300.5, 1.12 CPU seconds for r400.5 and 4.24 CPU seconds for r500.5.

For NuMVC, we set $\gamma = 0.5|V|$ and $\rho = 0.3$ for all runs, except for the challenging instance `frb100-40`, where $\gamma = 5000$ and $\rho = 0.3$. Note that there are also parameters in other state-of-the-art MVC (MC, MIS) algorithms, such as DLS-MC (Pullan & Hoos, 2006) and EWLS (Cai et al., 2010). Moreover, the parameters in DLS-MC and EWLS vary considerably on different instances. For each instance, each algorithm is performed 100 independent runs with different random seeds, where each run is terminated upon reaching a given cutoff time. The cutoff time is set to 2000 seconds for all instances except for the challenging instance `frb100-40`, for which the cutoff time is set to 4000 seconds due to its significant hardness.

For NuMVC, we report the following information for each instance:

- The optimal (or minimum known) vertex cover size ($VC^*$).

---

3. http://www.nlsde.buaa.edu.cn/~kexu/benchmarks/graph-benchmarks.htm
4. http://www.satcompetition.org
5. http://www.nlsde.buaa.edu.cn/~kexu/benchmarks/list-graph-papers.htm
6. http://www.shaoweicai.net/research.html
7. http://www.informatik.uni-freiburg.de/~srichter/
8. ftp://dimacs.rutgers.edu/pub/dsj/clique/

- The number of successful runs ("suc"). A run is said successful if a solution of size $VC^*$ is found.

- The "VC size" which shows the min (average, max) vertex cover size found by NuMVC in 100 runs.

- The averaged run time over all 100 runs ("time"). The run time of a successful run is the time to find the $VC^*$ solution, and that of a failed run is considered to be the cutoff time. For instances where NuMVC does not achieve a 100% success rate, we also report the averaged run time over only successful runs ("suc time"). The run time is measured in CPU seconds.

- The inter-quartile range (IQR) of the run time for 100 runs. The IQR is the difference between the 75th percentile and the 25th percentile of a sample. IQR is one of the most famous robust measures in data analysis (Hoaglin, Mosteller, & Tukey, 2000), and has been recommended as a measurement of closeness of the sampling distribution by the community of experimental algorithms (Bartz-Beielstein, Chiarandini, Paquete, & Preuss, 2010).

- The number of steps averaged over all 100 runs ("steps"). The steps of a successful run is those needed to find the $VC^*$ solution, while the steps of a failed run are those executed before the running is cut off. For instances where NuMVC does not achieve a 100% success rate, we also report the averaged steps over only successful runs ("suc steps").

If there are no successful runs for an instance, the "time" and "steps" columns are marked with "n/a". When the success rate of a solver on an instance is less than 75%, the 75th percentile of the run time sample is just the cutoff time and does not represent the real 75th percentile. In this case, we do not report the IQR, and instead we mark with "n/a" on the corresponding column. Actually, if the success rate of a solver on a certain instance is less than 75%, the solver should be considered not robust on that instance given the cutoff time.

### 6.3 Performance of NuMVC

In this section, we report a detailed performance of NuMVC on the two benchmarks.

#### 6.3.1 Performance of NuMVC on DIMACS Benchmark

The performance results of NuMVC on the DIMACS benchmark are displayed in Table 1. NuMVC finds optimal (or best known) solutions for 35 out of 37 DIMACS instances. Note that the 2 failed instances are both `brock` graphs. Furthermore, among the 35 successful instances, NuMVC does so consistently (i.e., in all 100 runs) for 32 instances, 24 of which are solved within 1 second. Overall, the NuMVC algorithm exhibits excellent performance on the DIMACS benchmark except for the `brock` graphs. Remark that the `brock` graphs are artificially designed to defeat greedy heuristics by explicitly incorporating low-degree vertices into the optimal vertex cover. Indeed, most algorithms preferring higher-degree vertices such as GRASP, RLS, k-opt, COVER and EWCC also failed in these graphs.

#### 6.3.2 Performance of NuMVC on BHOSLIB Benchmark

In Table 2, we illustrate the performance of NuMVC on the BHOSLIB benchmark. NuMVC successfully solves all BHOSLIB instances in terms of finding an optimal solution, and the size of the worst solution it finds never exceeds $VC^* + 1$. NuMVC finds optimal solutions with 100% success rate for 33 out of these 40 instances, and the averaged success rate over the remaining 7 instances is 82.57%. These results are dramatically better than existing results in the literature on this benchmark. Also, NuMVC finds a sub-optimal solution of size $VC^* + 1$ for all BSHOSLIB instances very quickly, always in less than 30 seconds. This indicates NuMVC can be used to approximate the MVC problem efficiently even under very limited time.

| Graph | | | NuMVC | | | |
|---|---|---|---|---|---|---|
| Instance | Vertices | $VC^*$ | suc | VC size | time(suc time) | steps(suc steps) |
| `brock200_2` | 200 | 188* | 100 | 188 | 0.126 | 137610 |
| `brock200_4` | 200 | 183* | 100 | 183 | 1.259 | 1705766 |
| `brock400_2` | 400 | 371* | 96 | 371(371.16,375) | 572.390(512.906) | 645631471(585032783) |
| `brock400_4` | 400 | 367* | 100 | 367 | 4.981 | 6322882 |
| `brock800_2` | 800 | 776* | 0 | 779 | n/a | n/a |
| `brock800_4` | 800 | 774* | 0 | 779 | n/a | n/a |
| `C125.9` | 125 | 91* | 100 | 91 | < 0.001 | 136 |
| `C250.9` | 250 | 206* | 100 | 206 | < 0.001 | 3256 |
| `C500.9` | 500 | 443* | 100 | 443 | 0.128 | 133595 |
| `C1000.9` | 1000 | 932 | 100 | 932 | 2.020 | 1154155 |
| `C2000.5` | 2000 | 1984 | 100 | 1984 | 2.935 | 231778 |
| `C2000.9` | 2000 | 1920 | 1 | 1920(1921.29,1922) | 1994.561(1393.303) | 777848959(564895994) |
| `C4000.5` | 4000 | 3982 | 100 | 3982 | 252.807 | 7802785 |
| `DSJC500.5` | 500 | 487* | 100 | 487 | 0.012 | 3800 |
| `DSJC1000.5` | 1000 | 985* | 100 | 985 | 0.615 | 134796 |
| `gen200_p0.9_44` | 200 | 156* | 100 | 156 | < 0.001 | 1695 |
| `gen200_p0.9_55` | 200 | 145* | 100 | 145 | < 0.001 | 69 |
| `gen400_p0.9_55` | 400 | 345* | 100 | 345 | 0.035 | 38398 |
| `gen400_p0.9_65` | 400 | 335* | 100 | 335 | < 0.001 | 1522 |
| `gen400_p0.9_75` | 400 | 325* | 100 | 325 | < 0.001 | 203 |
| `hamming8-4` | 256 | 240* | 100 | 240 | < 0.001 | 1 |
| `hamming10-4` | 1024 | 984* | 100 | 984 | 0.062 | 23853 |
| `keller4` | 171 | 160* | 100 | 160 | < 0.001 | 42 |
| `keller5` | 776 | 749* | 100 | 749 | 0.038 | 15269 |
| `keller6` | 3361 | 3302 | 100 | 3302 | 2.51 | 384026 |
| `MANN_a27` | 378 | 252* | 100 | 252 | < 0.001 | 6651 |
| `MANN_a45` | 1035 | 690* | 100 | 690 | 86.362 | 90642150 |
| `MANN_a81` | 3321 | 2221 | 27 | 2221(2221.94,2223) | 1657.880(732.897) | 571607432(251509010) |
| `p_hat300-1` | 300 | 292* | 100 | 292 | 0.003 | 100 |
| `p_hat300-2` | 300 | 275* | 100 | 275 | < 0.001 | 98 |
| `p_hat300-3` | 300 | 264* | 100 | 264 | 0.001 | 1863 |
| `p_hat700-1` | 700 | 689* | 100 | 689 | 0.011 | 1248 |
| `p_hat700-2` | 700 | 656* | 100 | 656 | 0.006 | 1103 |
| `p_hat700-3` | 700 | 638* | 100 | 638 | 0.008 | 2868 |
| `p_hat1500-1` | 1500 | 1488* | 100 | 1488 | 3.751 | 445830 |
| `p_hat1500-2` | 1500 | 1435* | 100 | 1435 | 0.071 | 5280 |
| `p_hat1500-3` | 1500 | 1406 | 100 | 1406 | 0.060 | 10668 |

Table 1: NuMVC performance results, averaged over 100 independent runs, for the DIMACS benchmark instances. The VC* column marked with an asterisk means that the minimum known vertex cover size has been proved optimal.

Besides the 40 BHOSLIB instances in Table 2, there is a challenging instance `frb100-40`, which has a hidden minimum vertex cover of size 3900. The designer of the BHOSLIB benchmark conjectured that this instance will not be solved on a PC in less than a day within the next two decades[9]. The latest record for this challenging instance is a 3902-sized vertex cover found by EWLS, and also EWCC.

We run NuMVC 100 independent trials within 4000 seconds on `frb100-40`, with $\gamma = 5000$ and $\rho = 0.3$ (this parameter setting yields the best performance among all combinations from $\gamma = 2000, 3000, ..., 6000$ and $\rho = 0.1, 0.2, ..., 0.5$). Among these 100 runs, 4 runs find a 3902-sized solution with the averaged time of 2955 seconds, and 93 runs find a 3903-sized solution (including 3902-sized) with the averaged time of 1473 seconds. Also, it is interesting to note that NuMVC can locate a rather good approximate solution for this hard instance very quickly: the size of vertex covers that NuMVC finds within 100 seconds is between 3903 and 3905.

Generally, finding a (k+1)-vertex cover is much easier than a $k$-vertex cover. Hence, for NuMVC, as well as most other MVC local search algorithms which also solve the MVC problem by solving the $k$-vertex cover problem iteratively, the majority of running time is used in finding the best vertex cover $C^*$ (of the run), and in trying, without success, to find a vertex cover of size $(|C^*| - 1)$.

### 6.4 Comparison with Other Heuristic Algorithms

In the recent literature there are five leading heuristic algorithms for MVC (MC, MIS), including three MVC algorithms COVER (Richter et al., 2007), EWLS (Cai et al., 2010) and EWCC (Cai et al., 2011), and two MC algorithms DLS-MC (Pullan & Hoos, 2006) and PLS (Pullan, 2006). Note that EWCC and PLS are the improved versions of EWLS and DLS-MC respectively, and show better performance over their original versions on DIMACS and BHOSLIB benchmarks. Therefore, we compare NuMVC only with PLS, COVER and EWCC.

When comparing NuMVC with other heuristic algorithms, we report $VC^*$, "suc", "time" as well as IQR. The averaged run time over only successful runs ("suc time") cannot indicate comparative performance of algorithms correctly unless the evaluated algorithms have close success rates, and can be calculated by $\frac{"time" * 100 - cutoff * (100 - "suc")}{"suc"}$, so we do not report these statistics. The results in bold indicate the best performance for an instance.

#### 6.4.1 Comparative Results on DIMACS Benchmark

The comparative results on the DIMACS benchmark are shown in Table 3. Most DIMACS instances are so easy that they can be solved by all solvers with 100% success rate within 2 seconds, and thus are not reported in the table. Actually, the fact that the DIMACS benchmark has been reduced to 11 useful instances really emphasizes the need to make a new benchmark.

---

9. http://www.nlsde.buaa.edu.cn/~kexu/benchmarks/graph-benchmarks.htm

| Graph | | | NuMVC | | | |
|---|---|---|---|---|---|---|
| Instance | Vertices | $VC^*$ | suc | VC size | time (suc time) | steps (suc steps) |
| frb30-15-1 | 450 | 420 | 100 | 420 | 0.045 | 37963 |
| frb30-15-2 | 450 | 420 | 100 | 420 | 0.053 | 44632 |
| frb30-15-3 | 450 | 420 | 100 | 420 | 0.191 | 173708 |
| frb30-15-4 | 450 | 420 | 100 | 420 | 0.049 | 41189 |
| frb30-15-5 | 450 | 420 | 100 | 420 | 0.118 | 105468 |
| frb35-17-1 | 595 | 560 | 100 | 560 | 0.515 | 386287 |
| frb35-17-2 | 595 | 560 | 100 | 560 | 0.447 | 334255 |
| frb35-17-3 | 595 | 560 | 100 | 560 | 0.178 | 129279 |
| frb35-17-4 | 595 | 560 | 100 | 560 | 0.563 | 422638 |
| frb35-17-5 | 595 | 560 | 100 | 560 | 0.298 | 218800 |
| frb40-19-1 | 760 | 720 | 100 | 720 | 0.242 | 208115 |
| frb40-19-2 | 760 | 720 | 100 | 720 | 4.083 | 3679770 |
| frb40-19-3 | 760 | 720 | 100 | 720 | 1.076 | 959874 |
| frb40-19-4 | 760 | 720 | 100 | 720 | 2.757 | 2473081 |
| frb40-19-5 | 760 | 720 | 100 | 720 | 10.141 | 9142719 |
| frb45-21-1 | 945 | 900 | 100 | 900 | 2.708 | 2029588 |
| frb45-21-2 | 945 | 900 | 100 | 900 | 4.727 | 3605881 |
| frb45-21-3 | 945 | 900 | 100 | 900 | 13.777 | 10447444 |
| frb45-21-4 | 945 | 900 | 100 | 900 | 3.973 | 3000680 |
| frb45-21-5 | 945 | 900 | 100 | 900 | 10.661 | 8059236 |
| frb50-23-1 | 1150 | 1100 | 100 | 1100 | 38.143 | 24628019 |
| frb50-23-2 | 1150 | 1100 | 100 | 1100 | 176.589 | 113569606 |
| frb50-23-3 | 1150 | 1100 | 95 | 1100(1100.05,1101) | 606.165(532.805) | 386342329(343518242) |
| frb50-23-4 | 1150 | 1100 | 100 | 1100 | 7.89 | 5092072 |
| frb50-23-5 | 1150 | 1100 | 100 | 1100 | 19.529 | 12690957 |
| frb53-24-1 | 1272 | 1219 | 86 | 1219(1219.14,1220) | 895.006(715.123) | 514619149(416394360) |
| frb53-24-2 | 1272 | 1219 | 100 | 1219 | 205.352 | 117980833 |
| frb53-24-3 | 1272 | 1219 | 100 | 1219 | 51.227 | 29376406 |
| frb53-24-4 | 1272 | 1219 | 100 | 1219 | 266.871 | 152982736 |
| frb53-24-5 | 1272 | 1219 | 100 | 1219 | 39.893 | 22817023 |
| frb56-25-1 | 1400 | 1344 | 100 | 1344 | 470.682 | 259903023 |
| frb56-25-2 | 1400 | 1344 | 97 | 1344(1344.03,1345) | 658.961(617.485) | 350048132(326853745) |
| frb56-25-3 | 1400 | 1344 | 100 | 1344 | 121.298 | 67043078 |
| frb56-25-4 | 1400 | 1344 | 100 | 1344 | 49.446 | 26030031 |
| frb56-25-5 | 1400 | 1344 | 100 | 1344 | 26.761 | 14109165 |
| frb59-26-1 | 1534 | 1475 | 88 | 1475(1475.12,1476) | 843.304(687.845) | 440874471(350993718) |
| frb59-26-2 | 1534 | 1475 | 38 | 1475(1475.62,1476) | 1677.801(1160.020) | 875964146(592010913) |
| frb59-26-3 | 1534 | 1475 | 96 | 1475(1475.04,1476) | 644.831(580.032) | 325417225(295226277) |
| frb59-26-4 | 1534 | 1475 | 79 | 1475(1475.21,1476) | 1004.550(741.208) | 517521634(375976753) |
| frb59-26-5 | 1534 | 1475 | 100 | 1475 | 61.907 | 31682895 |

Table 2: NuMVC performance results, averaged over 100 independent runs, for the BHOSLIB benchmark instances. All these BHOSLIB instances have a hidden optimal vertex cover, whose size is shown in the VC* column.

As indicated in Table 3, NuMVC outperforms COVER and EWCC on all instances, and is competitive with and complementary to PLS. For the eight hard instances on which at least one solver fails to achieve a 100% success rate, PLS dominates on the `brock` graphs while NuMVC dominates on the others, including the two putatively hardest instances `C2000.9` and `MANN_a81` (Richter et al., 2007; Grosso et al., 2008; Cai et al., 2011), as well as `keller6` and `MANN_a45`.

| Graph Instance | $VC^*$ | PLS | | COVER | | EWCC | | NuMVC | |
|---|---|---|---|---|---|---|---|---|---|
| | | suc | time (IQR) | suc | time (IQR) | suc | time (IQR) | suc | time (IQR) |
| `brock400_2` | 371* | **100** | **0.15** (0.16) | 3 | 1947 (n/a) | 20 | 1778 (n/a) | 96 | 572 (646) |
| `brock400_4` | 367* | 100 | **0.03** (0.03) | 82 | 960 (988) | 100 | 25.38 (25.96) | 100 | 4.98 (6.14) |
| `brock800_2` | 776* | **100** | **3.89** (3.88) | 0 | n/a | 0 | n/a | 0 | n/a |
| `brock800_4` | 774* | **100** | **1.31** (1.52) | 0 | n/a | 0 | n/a | 0 | n/a |
| `C2000.9` | 1920 | 0 | n/a | 0 | n/a | 0 | n/a | **1** | **1994** (n/a) |
| `C4000.5` | 3982 | 100 | **67** (59) | 100 | 658 (290) | 100 | 739 (903) | 100 | 252 (97) |
| `gen400_p0.9_55` | 345* | 100 | 15.17 (17) | 100 | 0.35 (0.1) | 100 | 0.05 (0.04) | 100 | **0.03** (0.01) |
| `keller6` | 3302 | 92 | 559 (515) | 100 | 68 (6) | 100 | 3.76 (3.57) | **100** | **2.51** (0.76) |
| `MANN_a45` | 690* | 1 | 1990 (n/a) | 94 | 714 (774) | 88 | 763 (766) | **100** | **86** (95) |
| `MANN_a81` | 2221 | 0 | n/a | 1 | 1995 (n/a) | 1 | 1986 (n/a) | **27** | **1657** (n/a) |
| `p_hat1500-1` | 1488* | 100 | **2.36** (3.07) | 100 | 18.10 (17.23) | 100 | 9.79 (9.77) | 100 | 3.75 (3.19) |

Table 3: Comparison of NuMVC with other state-of-the-art heuristic algorithms on the DIMACS benchmark. The VC* column marked with an asterisk means that the minimum known vertex cover size has been proved optimal.

For `C2000.9`, only NuMVC finds a 1920-sized solution, and it also finds a 1921-sized solution in 70 runs, while this number is 31, 6 and 32 for PLS, COVER, and EWCC respectively. Note that PLS performs well on the `brock` family because it comprises three sub-algorithms, one of which favors the lower degree vertices.

Table 3 indicates that `C2000.9` and `MANN_a81` remain very difficult for modern algorithms, as none of the algorithms can solve them with a good success rate in reasonable time. On the other hand, other instances can be solved quickly (in less than 100 seconds) by at least one algorithm, PLS or NuMVC, with a low IQR value (always less than 100), which indicates quite stable performance.

### 6.4.2 Comparative Results on BHOSLIB Benchmark

In Table 4, we present comparative results on the BHOSLIB benchmark. For concentrating on the considerable gaps in comparisons, we do not report the results on the two groups of small instances (`frb30` and `frb35`), which can be solved within several seconds by all solvers.

The results in Table 4 illustrate that NuMVC significantly outperforms the other algorithms on all BHOSLIB instances, in terms of both success rate and averaged run time, which are also demonstrated in Figure 1. We take a further look at the comparison between NuMVC and EWCC, as EWCC performs obviously better than PLS and COVER on this benchmark. NuMVC solves 33 instances out of 40 with 100% success rate, 4 more instances than EWCC does. For those instances solved by both algorithms with 100% success rate, the overall averaged run time is 25 seconds for NuMVC and 74 seconds for EWCC. For other instances, the averaged success rate is 90% for NuMVC, compared to 50% for EWCC.

The excellent performance of NuMVC is further underlined by the large gaps between NuMVC and the other solvers on the hard instances. For example, on the instances where all solvers fail to find an optimal solution with 100% success rate, NuMVC achieves an overall averaged success rate of 82.57%, dramatically better than those of PLS, COVER and EWCC, which are 0.85%, 17.43% and 35.71% respectively. Obviously, the experimental results show that NuMVC delivers the best performance for this hard random benchmark, vastly improving the existing performance results. We also observe that, NuMVC always has the minimum IQR value for all instances, which indicates that apart from its efficiency, the robustness of NuMVC is also better than other solvers.

| Graph Instance | $VC^*$ | PLS | | COVER | | EWCC | | NuMVC | |
|---|---|---|---|---|---|---|---|---|---|
| | | suc | time (IQR) | suc | time (IQR) | suc | time (IQR) | suc | time (IQR) |
| frb40-19-1 | 720 | 100 | 10.42 (10.38) | 100 | 1.58 (0.55) | 100 | 0.55 (0.48) | 100 | **0.24** (0.18) |
| frb40-19-2 | 720 | 100 | 85.25 (72.75) | 100 | 17.18 (16.09) | 100 | 11.30 (14.21) | 100 | **4.08** (3.77) |
| frb40-19-3 | 720 | 100 | 9.06 (10.21) | 100 | 5.06 (4) | 100 | 2.97 (2.35) | 100 | **1.07** (1.03) |
| frb40-19-4 | 720 | 100 | 77.39 (90.56) | 100 | 11.79 (8.67) | 100 | 13.79 (16.05) | 100 | **2.76** (2.83) |
| frb40-19-5 | 720 | 95 | 496 (529.25) | 100 | 124 (131) | 100 | 41.71 (39.08) | 100 | **10.14** (10.54) |
| frb45-21-1 | 900 | 100 | 52.31 (55.5) | 100 | 14.34 (12.8) | 100 | 9.07 (9.3) | 100 | **2.71** (2.6) |
| frb45-21-2 | 900 | 100 | 170 (202.2) | 100 | 38 (35.4) | 100 | 15 (14.1) | 100 | **5** (5.1) |
| frb45-21-3 | 900 | 21 | 1737 (n/a) | 100 | 110 (121) | 100 | 56 (70.4) | 100 | **14** (11.9) |
| frb45-21-4 | 900 | 100 | 111 (130) | 100 | 21 (18) | 100 | 15 (12.5) | 100 | **4** (4.3) |
| frb45-21-5 | 900 | 100 | 261 (300) | 100 | 105 (103 ) | 100 | 42 (40.1) | 100 | **11** (10.9) |
| frb50-23-1 | 1100 | 30 | 1658 (640) | 100 | 268 (305) | 100 | 124 (135) | 100 | **38** (46) |
| frb50-23-2 | 1100 | 3 | 1956 (n/a) | 48 | 1325 (n/a) | 82 | 905 (1379) | 100 | **177** (149) |
| frb50-23-3 | 1100 | 2 | 1989 (n/a) | 39 | 1486 (n/a) | 56 | 1348 (n/a) | 95 | **606** (788) |
| frb50-23-4 | 1100 | 100 | 93 (80) | 100 | 33 (25) | 100 | 24 (27) | 100 | **8** (7) |
| frb50-23-5 | 1100 | 79 | 967 (1305) | 100 | 168 (246) | 100 | 85 (97) | 100 | **19** (19) |
| frb53-24-1 | 1219 | 1 | 1982 (n/a) | 17 | 1796 (n/a) | 30 | 1696 (n/a) | 86 | **895** (1099) |
| frb53-24-2 | 1219 | 6 | 1959 (n/a) | 50 | 1279 (n/a) | 81 | 1006 (1270) | 100 | **205** (200) |
| frb53-24-3 | 1219 | 20 | 1771 (n/a) | 99 | 273 (223) | 100 | 117 (136) | 100 | **51** (48) |
| frb53-24-4 | 1219 | 21 | 1782 (n/a) | 48 | 1428 (n/a) | 81 | 900 (1480) | 100 | **266** (311) |
| frb53-24-5 | 1219 | 10 | 1955 (n/a) | 95 | 423 (315) | 100 | 125 (115) | 100 | **40** (44) |
| frb56-25-1 | 1344 | 1 | 1993 (n/a) | 24 | 1698 (n/a) | 56 | 1268 (n/a) | 100 | **470** (466) |
| frb56-25-2 | 1344 | 0 | n/a | 17 | 1598 (n/a) | 52 | 1387 (n/a) | 97 | **659** (780) |
| frb56-25-3 | 1344 | 0 | n/a | 97 | 537 (692) | 100 | 285 (250) | 100 | **121** (118) |
| frb56-25-4 | 1344 | 11 | 1915 (n/a) | 93 | 476 (460) | 100 | 183 (188) | 100 | **50** (49) |
| frb56-25-5 | 1344 | 27 | 1719 (n/a) | 100 | 168 (128) | 100 | 80 (81) | 100 | **27** (23) |
| frb59-26-1 | 1475 | 0 | n/a | 16 | 1607 (n/a) | 21 | 1778 (n/a) | 88 | **843** (849) |
| frb59-26-2 | 1475 | 0 | n/a | 9 | 1881 (n/a) | 7 | 1930 (n/a) | 37 | **1677** (n/a) |
| frb59-26-3 | 1475 | 3 | 1978 (n/a) | 21 | 1768 (n/a) | 64 | 1294 (n/a) | 96 | **636** (788) |
| frb59-26-4 | 1475 | 0 | n/a | 3 | 1980 (n/a) | 20 | 1745 (n/a) | 79 | **1004** (1391) |
| frb59-26-5 | 1475 | 30 | 1708 (420) | 98 | 431 (476) | 100 | 174 (182) | 100 | **62** (70) |

Table 4: Comparison of NuMVC with other state-of-the-art local search algorithms on the BHOSLIB benchmark. All these BHOSLIB instances have a hidden optimal vertex cover, whose size is shown in the VC* column.

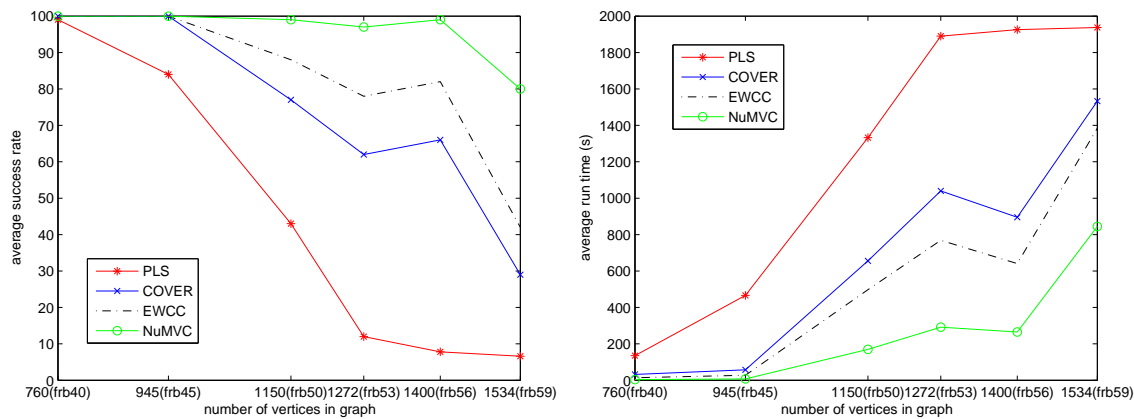

Figure 1: Comparison of NuMVC and other local search algorithms on the BHOSLIB benchmark in terms of success rate (left) and averaged run time (right)

We also compare NuMVC with COVER and EWCC on the challenging instance `frb100-40`. Given the failure of PLS on large BHOSLIB instances, we do not run PLS on this instance. The comparative results on `frb100-40` are shown in Table 5, which indicates that NuMVC significantly outperforms COVER and EWCC on this challenging instance.

Finally, we would like to remark that the performance of NuMVC on the BHOSLIB benchmark is better than a four-core version of CLS (Pullan et al., 2011), even if we do not divide the run time of NuMVC by 4 (the number of cores utilized by CLS). If we consider the machine speed ratio and divide the run time of NuMVC by 4, then NuMVC would be dramatically better than CLS on the BHOSLIB benchmark.

| Size of VC | COVER | | EWCC | | NuMVC | |
|---|---|---|---|---|---|---|
| | suc | avg suc time | suc | avg suc time | suc | avg suc time |
| 3902 | 0 | n/a | 1 | 2586 | **4** | **2955** |
| ≤ 3903 | 33 | 2768 | 79 | 2025 | **93** | **1473** |

Table 5: Comparative results on the `frb100-40` challenging instance. Each solver is executed 100 times on this instance with a timeout of 4000 seconds.

## 6.5 Comparison with Exact Algorithms

In this section, we compare NuMVC with a state-of-the-art exact Maximum Clique algorithm. Generally, exact algorithms and heuristic algorithms are somewhat complementary in their applications. Usually, exact algorithms find solutions for structured instances faster while heuristic algorithms are faster on random ones.

Compared to MVC and MIS, many more exact algorithms are designed for the Maximum Clique problem (Carraghan & Pardalos, 1990; Fahle, 2002; Östergård, 2002; Régin, 2003; Tomita & Kameda, 2009; Li & Quan, 2010b, 2010a). The recent branch-and-bound MC algorithm MaxCLQ (Li & Quan, 2010b) which utilizes MaxSAT inference technologies (Li, Manyà, & Planes, 2007) to improve upper bounds shows considerable progress. Experimental results of MaxCLQ (Li & Quan, 2010b) on some random graphs and DIMACS instances indicate that MaxCLQ significantly outperforms previous exact MC algorithms. The MaxCLQ algorithm is further improved using two strategies called Extended Failed Literal Detection and Soft Clause Relaxation, resulting in a better algorithm denoted by MaxCLQdyn+EFL+SCR (Li & Quan, 2010a). Due to the great success of MaxCLQdyn+EFL+SCR, we compare our algorithm only with MaxCLQdyn+EFL+SCR.

We compare NuMVC with MaxCLQdyn+EFL+SCR on the DIMACS benchmark instances. The results of MaxCLQdyn+EFL+SCR are taken from the previous work (Li & Quan, 2010a). MaxCLQdyn+EFL+SCR is not evaluated on the BHOSLIB benchmark which is much harder and requires more effective technologies for exact algorithms (Li & Quan, 2010a).

The run time results of MaxCLQdyn+EFL+SCR are obtained on a 3.33 GHz Intel Core 2 Duo CPU with linux and 4 Gb memory, which required 0.172 seconds for r300.5, 1.016 seconds for r400.5 and 3.872 seconds for r500.5 to execute the DIMACS machine benchmarks (Li & Quan, 2010a). The corresponding run time for our machine is 0.19, 1.12 and 4.24 seconds. So, we multiply the reported run time of MaxCLQdyn+EFL+SCR by 1.098 (=(4.24/3.872+1.12/1.016)/2=1.098, the average of the two largest ratios). This normalization is based on the methodology established in the Second DIMACS Implementation Challenge for Cliques, Coloring, and Satisfiability, and is widely used for comparing different MaxClique algorithms (Pullan & Hoos, 2006; Pullan, 2006; Li & Quan, 2010b, 2010a).

| Graph | | NuMVC | | MaxCLQdyn+EFL | Graph | | NuMVC | | MaxCLQdyn+EFL |
|---|---|---|---|---|---|---|---|---|---|
| Instance | $VC^*$ | suc | time | +SCR time | Instance | $VC^*$ | suc | time | +SCR time |
| brock400_2 | 371 | 96 | 572.39 | **125.06** | p_hat300-3 | 264 | 100 | **0.001** | 1.31 |
| brock400_3 | 369 | 100 | **8.25** | 251.44 | p_hat700-2 | 656 | 100 | **0.006** | 3.27 |
| brock400_4 | 367 | 100 | **4.98** | 119.24 | p_hat700-3 | 638 | 100 | **0.008** | 1141.92 |
| brock800_2 | 776 | 0 | n/a | **5138.10** | p_hat1000-2 | 954 | 100 | **0.019** | 108.94 |
| brock800_3 | 775 | 0 | n/a | **3298.39** | p_hat1000-3 | 932 | 100 | **0.032** | 113860.40 |
| brock800_4 | 774 | 0 | n/a | **2391.44** | p_hat1500-1 | 1488 | 100 | 3.75 | **3.10** |
| keller5 | 749 | 100 | **0.04** | 6884.46 | p_hat1500-2 | 1435 | 100 | **0.071** | 866.51 |
| MANN_a27 | 252 | 100 | **<0.001** | 0.17 | sanr200_0.9 | 158 | 100 | **<0.001** | 5.20 |
| MANN_a45 | 690 | 100 | 86.86 | **21.169** | sanr400_0.7 | 379 | 100 | **0.008** | 97.72 |

Table 6: Comparison of NuMVC with the state-of-the-art exact MaxClique algorithm MaxCLQ-dyn+EFL+SCR for the DIMACS benchmark.

In Table 6, we present the performance of NuMVC and MaxCLQdyn+EFL+SCR on the DIMACS instances. The results indicate that NuMVC finds an optimal solution much faster than MaxCLQdyn+EFL+SCR on random instances such as the p_hat and sanr instances. We believe that similar results would hold for other hard random benchmarks like BHOSLIB ones, as MaxCLQdyn+EFL+SCR is not evaluated on these instances due to their high hardness (Li & Quan, 2010a), while NuMVC performs very well on them.

For structured instances, we note that MaxCLQdyn+EFL+SCR is mainly evaluated on the brock instances where NuMVC performs worst, but not on the open DIMACS instances such as `MANN_a81`, `johnson32-2-4` and `keller6`, which remain very difficult to solve by exact algorithms (Li & Quan, 2010a). Although MaxCLQdyn+EFL+SCR overall performs better, NuMVC also finds an optimal solution significantly faster than MaxCLQdyn+EFL+SCR on some structured instances, such as the two `brock` instances and `keller5`.

Finally, we would like to note that although heuristic solvers can find optimal solutions fast, they are unable to prove the optimality of the solutions they find. On the other hand, the run time of an exact algorithm is spent not only on finding an optimal solution but also on proving its optimality. In this sense, heuristic and exact algorithms cannot be compared in a fair way. Nevertheless, our experiments suggest that heuristic approaches are appealing for solving large instances in reasonable short time.

## 7. Discussions

In this section, we first explore the run-time distribution of NuMVC on some representative instances, and then investigate the effectiveness of the two-stage exchange strategy and the forgetting mechanism in NuMVC. Finally, we analyze the performance of NuMVC with different settings to its two parameters for the forgetting mechanism, which shows that NuMVC is not sensitive to the parameters.

### 7.1 Run-time Distributions of NuMVC

In this subsection, we conduct an empirical study to gain deeper insights of the run-time behavior of NuMVC. More specifically, we study the run-time distribution of NuMVC on several representative instances. For the purpose of comparison, we also report the run-time distribution of EWCC, which is the best competing MVC local search solver.

Consider a randomized algorithm solving a given optimization problem instance, and halting as soon as an optimal solution is found. The run time of the algorithm can be viewed as a random variable, which is fully described by its distribution, commonly referred to as the run-time distribution (RTD) in the literature about algorithm performance modeling (Hoos & Stützle, 2004; Bartz-Beielstein et al., 2010). The methodology of studying the run-time behavior of algorithms based on RTDs has been widely used in empirical analysis of heuristic algorithms (Hoos & Stützle, 1999; Finkelstein, Markovitch, & Rivlin, 2003; Watson, Whitley, & Howe, 2005; Pullan & Hoos, 2006). We also follow the same methodology in our study here.

For studying typical run-time behaviour, we choose instances where NuMVC reaches an optimal solution in all 100 runs, and are of appropriate difficulty. For the DIMACS benchmark, we select `brock400_4` and `MANN_a45`, both of which are of reasonable size and hardness. Also, these two instances represent two typical instance classes for NuMVC, as NuMVC has poor performance on the `brock` instances, while it dominates other heuristic algorithms on the `MANN` instances. For the BHOSLIB benchmark, `frb56-25-5` and `frb59-26-5` are selected. These are appropriate instances for studying the run-time behavior of NuMVC, since they are neither too easy that can be solved in a short time nor too difficult to reach a 100% success rate.

The empirical RTD graphs of NuMVC and EWCC are shown in Figure 2 (the RTD for each instance is based on 100 independent runs that all reach a respective optimal solution). According to the graphs, NuMVC shows a large variability in run time. Further investigation indicates that these RTDs are quite well approximated by exponential distributions, labeled $ed[m](x) = 1 - 2^{-x/m}$, where $m$ is the median of the distribution. To test the goodness of the approximations, we use a

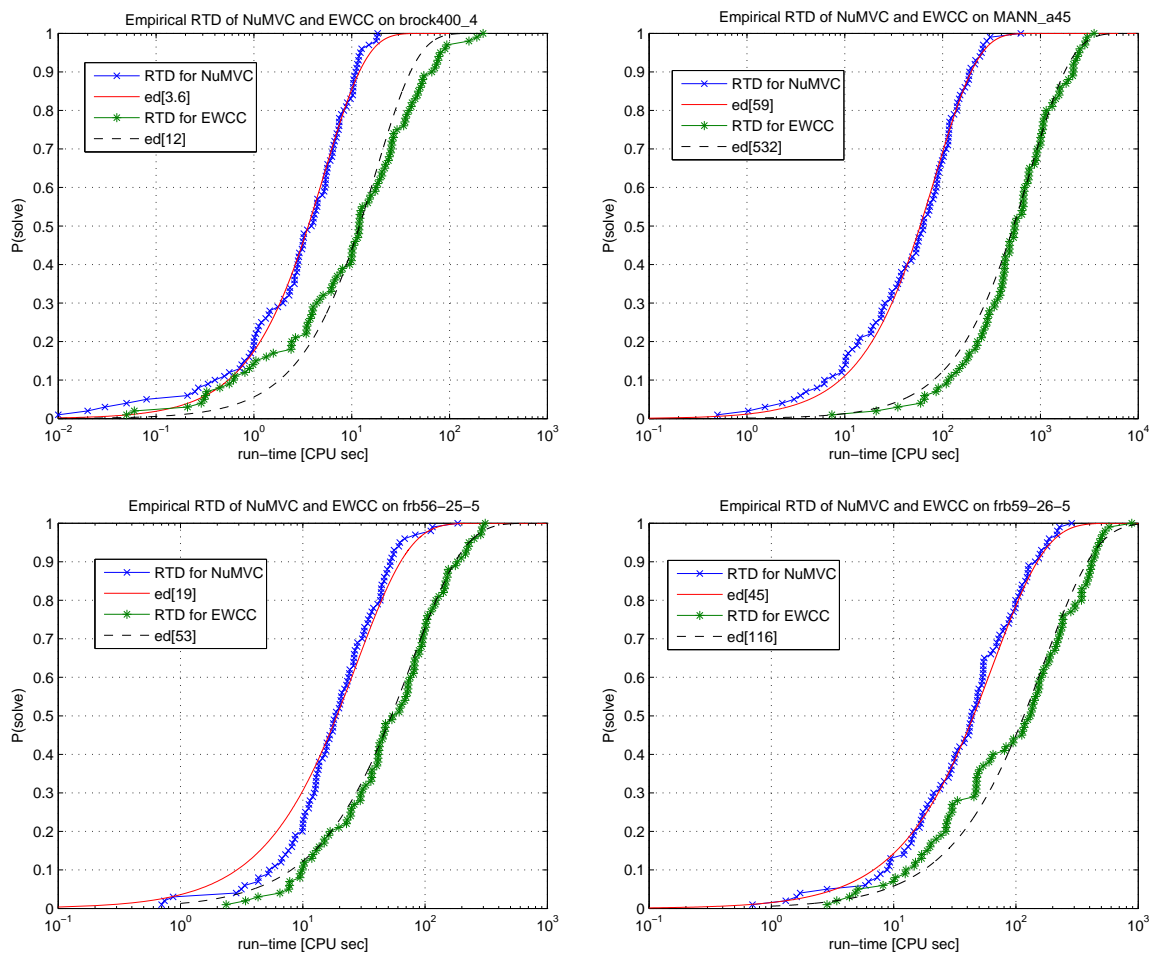

Figure 2: Run-time distributions (RTDs) of NuMVC and EWCC applied to two DIMACS instances (top) and two BHOSLIB instances (bottom); these empirical RTDs are well approximated by exponential distributions, labeled $ed[m](x) = 1 - 2^{-x/m}$ in the plots.

*Kolmogorov-Smirnov* test, which fails to reject the null hypothesis that the sampled run time stems from the exponential distributions shown in the figures at a standard confidence level of $\alpha = 0.05$ with *p-values* between 0.19 and 0.88. For EWCC, the *Kolmogorov-Smirnov* test shows its RTDs on `MANN_a45` and the two BHOSLIB instances are also exponential distributions, while its RTD on `brock400_4` is not from an exponential distribution.

The observation of exponential RTDs of NuMVC is consistent with similar results for other high performance SLS algorithms, e.g., for MaxClique (Pullan & Hoos, 2006), for SAT (Hoos & Stützle, 1999), for MAXSAT (Smyth, Hoos, & Stützle, 2003), and for scheduling problems (Watson et al., 2005). By the arguments (Hoos & Stützle, 1999; Hoos & Stützle, 2004) made for stochastic local search algorithms characterized by an exponential RTD, we conclude that, for NuMVC, the probability of finding an optimal solution within a fixed amount of time (or steps) does not depend on the run time in the past. Consequently, it is very robust *w.r.t.* the cutoff time and thus, the restart time. Therefore, performing multiple independent runs of NuMVC in parallel will result in close-to-optimal parallelization speedup. Similar observations were made for most of the other DIMACS instances and BHOSLIB instances.

Of practical interest is also the RTD analysis for NuMVC on difficult instances for which all algorithms in our experiments fail to achieve a high success rate (i.e., 40%). The RTDs in these cases would show where the algorithm stagnates and suggest an a-posteriori restart time for the algorithm. For this purpose, we select `MANN_a81` and `frb59-26-2` for analysis. The RTDs of NuMVC on these two instances are illustrated in Figure 3. Interestingly, from these RTDs we do not observe any obvious stagnation, which again confirms that NuMVC is robust $w.r.t.$ the cutoff time and thus the restart time. Therefore, by increasing the cutoff time, we can expect a higher success rate of the algorithm on these difficult instances.

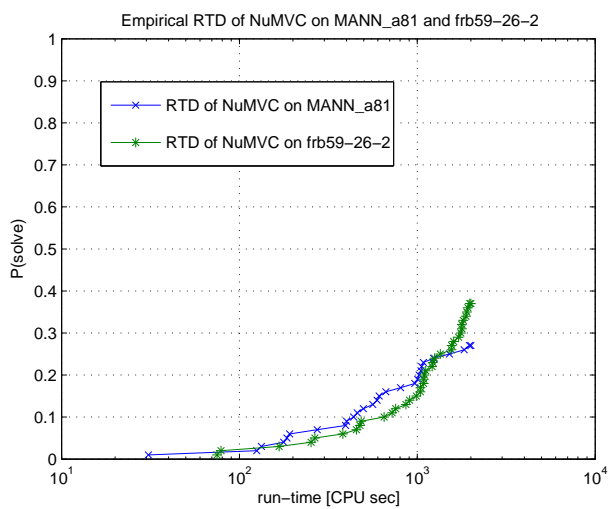

Figure 3: Run-time distributions (RTDs) of NuMVC on `MANN_a81` and `frb59-26-2` instances, for which NuMVC finds an optimal (or best known) solution in less than half runs.

## 7.2 Effectiveness of Two-Stage Exchange

To study the effectiveness of the two-stage exchange strategy, we compare NuMVC with its alternative algorithm $NuMVC_0$ which selects two vertices for exchanging simultaneously. In each step, $NuMVC_0$ first chooses an uncovered edge $e$ uniformly at random, and then evaluates each pair of vertices $u$ and $v$ where $u$ is in the current candidate solution and $v$ is one endpoint of $e$ such that $confChange(v) = 1$. For evaluating the benefit (i.e., the decrement of the $cost$ function) of exchanging a vertex pair $u$ and $v$, $NuMVC_0$ first checks whether they are neighbors. If $u$ and $v$ are neighbors, the benefit is $dscore(u) + dscore(v) + w(e\{u, v\})$; otherwise, the benefit is $dscore(u) + dscore(v)$. $NuMVC_0$ selects the vertex pair with the greatest benefit to exchange.

In the NuMVC (and also $NuMVC_0$) algorithm, there are only two candidate vertices to add to the current candidate solution $C$ (i.e., the endpoints of the selected uncovered edge). Hence, in the worst case, NuMVC performs $2 + |C|$ evaluations, while $NuMVC_0$ has to evaluate $2 \times |C|$ pairs of vertices. Moreover, NuMVC only needs to check the $dscore$ of a vertex in each (vertex) evaluation, while $NuMVC_0$ performs a vertex-pair evaluation which involves a pair of vertices and their relationship, and thus is more time-consuming. Based on the above analysis, we conjecture that the complexity per step of NuMVC is at least 2 times lower than that of $NuMVC_0$. Also, as we have mentioned in Section 3, the two-stage exchange strategy is less greedy than the one selecting two vertices for exchanging simultaneously, as $NuMVC_0$ does.

The investigation is carried out on 4 DIMACS instances from different families as well as 12 BHOSLIB instances. For the DIMACS benchmark, we select `brock400_2`, `C4000.5`, `MANN_a45`, and `p_hat_1500-1`. These instances have different characteristics, as described below (Pullan et al., 2011). Note that the following conclusions on DIMACS instances are for the complementary DIMACS graphs.

- The DIMACS `brock` instances have minimum vertex covers that consist of medium to lower degree vertices, and are designed to defeat greedy heuristics.

- The DIMACS `C` and `p_hat_1500-1` instances have minimum vertex covers that consist of higher degree vertices and can be effectively solved by greedy heuristics.

- The DIMACS `MANN` instances have a large proportion of plateaus in the instance search-space, and thus greedy heuristics are unsuitable to solve them.

- The BHOSLIB instances have minimum vertex covers consisting of vertices whose distribution of vertex degree closely matches that for the complete graph. These are difficult instances for both greedy and diversification heuristics.

| Graph Instance | $VC^*$ | NuMVC | | | | $NuMVC_0$ | | | |
|---|---|---|---|---|---|---|---|---|---|
| | | suc | time | steps | #steps/sec $(10^5)$ | suc | time | steps | #steps/sec $(10^5)$ |
| `brock400_2` | 371 | **96** | **572** | **645631471** | 11.3 | 19 | 1861 | 837844749 | 4.5 |
| `C4000.5` | 3982 | 100 | **252** | 7802785 | 0.3 | 100 | 607 | **6343304** | 0.1 |
| `MANN_a45` | 690 | 100 | **86** | **90642150** | 10.5 | 100 | 564 | 186350533 | 3.3 |
| `p_hat_1500-1` | 1488 | 100 | **3.75** | 445830 | 1.2 | 100 | 13.24 | **381762** | 0.3 |
| `frb50-23-1` | 1100 | 100 | **38** | 24628019 | 6.5 | 100 | 88 | **18125042** | 2.1 |
| `frb50-23-2` | 1100 | 100 | **177** | 113569606 | 6.4 | 100 | 499 | **104841043** | 2.1 |
| `frb50-23-3` | 1100 | 95 | **606** | 386342329 | 6.4 | 63 | 1312 | 262559614 | 2.0 |
| `frb53-24-1` | 1219 | **86** | **895** | 514619149 | 5.7 | 45 | 1595 | 286396840 | 1.8 |
| `frb53-24-2` | 1219 | 100 | **205** | 117980833 | 5.8 | 100 | 557 | **105863802** | 1.9 |
| `frb53-24-3` | 1219 | 100 | **51** | 29376406 | 5.8 | 100 | 106 | **19685358** | 1.9 |
| `frb56-25-1` | 1344 | **100** | **470** | 259903023 | 5.5 | 72 | 1088 | 184323492 | 1.7 |
| `frb56-25-2` | 1344 | 97 | **659** | 350048132 | 5.3 | 52 | 1499 | 254973016 | 1.7 |
| `frb56-25-3` | 1344 | 100 | **121** | 67043078 | 5.5 | 100 | 253 | **43062419** | 1.7 |
| `frb59-26-1` | 1475 | **88** | **843** | 440874471 | 5.2 | 45 | 1572 | 251520339 | 1.6 |
| `frb59-26-2` | 1475 | **37** | **1677** | 875964146 | 5.2 | 21 | 1853 | 315425608 | 1.7 |
| `frb59-26-3` | 1475 | **96** | **636** | 325417225 | 5.1 | 69 | 1545 | 247273810 | 1.6 |

Table 7: Comparative performance of NuMVC and $NuMVC_0$ which selects two vertices for exchanging simultaneously. The results are based on 100 independent runs for each solver on each instance.

The comparative results of NuMVC and $NuMVC_0$ are presented in Table 7. The results show that NuMVC significantly outperforms $NuMVC_0$ in terms of averaged run time, primarily due to its much lower complexity per step. In each second, NuMVC performs 3-4 times more steps than NuMVC$_0$, which supports our conjecture that the complexity per step of NuMVC is more than 2 times lower than that of NuMVC$_0$.

Now we turn our attention to comparing NuMVC and NuMVC$_0$ in terms of step performance, which is independent from the complexity per step. For `brock` and `MANN` graphs which are difficult for greedy heuristics, NuMVC has a significantly better step performance than NuMVC$_0$. On the other hand, for greedy-friendly graphs such as `C4000.5` and `p_hat_1500-1`, NuMVC needs more steps to converge to an optimal solution than NuMVC$_0$ does. These observations support our argument that the two-stage exchange strategy is less greedy than the one that selects two vertices for exchanging simultaneously, as NuMVC$_0$ does.

We also observe that the step performance of NuMVC$_0$ is better than that of NuMVC on BHOSLIB instances. For instance, on those BHOSLIB instances where both algorithms have a 100% success rate, NuMVC needs about 1.2 times more steps than NuMVC$_0$ to find an optimal solution. This is what we do not expect and cannot yet explain. Nevertheless, as NuMVC makes rather rapid modifications to a solution, a little degrade in step performance does not hurt.

| Graph | PLS | COVER | EWCC | NuMVC |
| Instance | #steps/sec | #steps/sec | #steps/sec | #steps/sec |
| --- | --- | --- | --- | --- |
| `C4000.5` | 85,318 | 8,699 | 11,927 | 30,963 |
| `MANN_a45` | 1,546,625 | 279,514 | 578,656 | 1,053,978 |
| `p_hat_1500-1` | 170,511 | 19,473 | 34,111 | 118,888 |
| `frb53-24-5` | 841,346 | 128,971 | 219,038 | 570,425 |
| `frb56-25-5` | 801,282 | 116,618 | 199,441 | 522,561 |
| `frb59-26-5` | 706,436 | 108,534 | 189,536 | 511,014 |

Table 8: Complexity per step on selected instances

To further demonstrate the low complexity per step of NuMVC, we compare the number of search steps per second between NuMVC and other state-of-the-art heuristic solvers on representative instances. As indicated in Table 8, NuMVC executes many more steps in each second than the other two MVC local search solvers COVER and EWCC do. For the instances in Table 8, each second NuMVC executes 4-6 times more steps than COVER, and 3-4 times more steps than EWCC. This indicates that the two-stage exchange strategy can significantly accelerate MVC local search algorithms. Although PLS performs more steps per second than NuMVC, it is an MC local search algorithm whose search scheme is essentially different from those of MVC local search algorithms.

### 7.3 Effectiveness of the Forgetting Mechanism

To study the effectiveness of the forgetting mechanism in NuMVC, we compare NuMVC with its two alternative algorithms NuMVC$_1$ and NuMVC$_2$, which are obtained from NuMVC by modifying the edge weighting scheme as below.

- NuMVC$_1$ works in the same way as NuMVC, except for not using the forgetting mechanism, that is, deleting line 18 from Algorithm 1.

- NuMVC$_2$ adopts the forgetting mechanism used in DLS-MC (Pullan & Hoos, 2006) for the weighting scheme. More specifically, NuMVC$_2$ increases all weights of uncovered edges by one at the end of each step, and performs a forgetting operation every $pd$ steps by decreasing weights by one for all edges whose weights are greater than one. Note that $pd$ is an instance-dependent parameter.

The experiments were carried out with some representative instances from both benchmarks. For the DIMACS benchmark, we select `brock400_2`, `C4000.5`, `keller6`, and `MANN_a45`, which are from different classes and of appropriate difficulty. For the BHOSLIB benchmark, we select three instances for each of the three largest-sized instance groups respectively.

| Graph | | | NuMVC | | NuMVC$_1$ | | NuMVC$_2$ | | |
|---|---|---|---|---|---|---|---|---|---|
| Instance | Vertices | $VC^*$ | suc | time | suc | time | $pd$ ($10^2$) | suc | time |
| `brock400_2` | 400 | 371 | 96 | 572 | 22 | 1781 | 15 | **100** | **21** |
| `C4000.5` | 4000 | 3982 | 100 | **252** | 100 | 270 | 60 | 100 | 327 |
| `keller6` | 3361 | 3302 | 100 | **2.51** | 100 | 2.95 | 750 | 100 | 4.26 |
| `MANN_a45` | 1035 | 690 | **100** | **86** | 65 | 1187 | 8 | 100 | 113 |
| `frb53-24-1` | 1272 | 1219 | **86** | **895** | 60 | 925 | 100 | 78 | 901 |
| `frb53-24-2` | 1272 | 1219 | 100 | 205 | 100 | 243 | 100 | 100 | **201** |
| `frb53-24-3` | 1272 | 1219 | 100 | 51 | 100 | **49** | 100 | 100 | 52 |
| `frb56-25-1` | 1400 | 1344 | **100** | **470** | 85 | 914 | 130 | 91 | 595 |
| `frb56-25-2` | 1400 | 1344 | **97** | **659** | 63 | 1209 | 130 | 81 | 739 |
| `frb56-25-3` | 1400 | 1344 | 100 | 121 | 100 | **111** | 130 | 100 | 117 |
| `frb59-26-1` | 1534 | 1475 | 88 | **843** | 64 | 1229 | 150 | 85 | 907 |
| `frb59-26-2` | 1534 | 1475 | 37 | 1677 | 21 | 1894 | 150 | **45** | **1439** |
| `frb59-26-3` | 1534 | 1475 | 96 | 636 | 83 | 997 | 150 | **97** | 652 |

Table 9: Comparative performance of NuMVC and its two alternatives NuMVC$_1$ and NuMVC$_2$. Each algorithm is performed 100 times on each instance.

An apparent observation from Table 9 is that the two algorithms with a forgetting mechanisms (i.e., NuMVC and NuMVC$_2$) outperform NuMVC$_1$ on almost all instances. Particularly, due to the missing of a forgetting mechanism, NuMVC$_1$ performs significantly worse than the other two algorithms on `brock` and `MANN` graphs. On the other hand, Table 9 demonstrates that NuMVC and NuMVC$_2$ exhibit competitive performance on the BHOSLIB benchmark, and dominate on different types of DIMACS instances. More specifically, NuMVC outperforms NuMVC$_2$ on `C4000.5`, `keller6` and `MANN_a45`, but performs significantly worse than NuMVC$_2$ on `brock400_2`. In order to find out the genuine performance of NuMVC$_2$ on `brock` instances, we test NuMVC$_2$ on the larger `brock800_2` and `brock800_4` instances. The results show that these two large `brock` instances are substantially more difficult than the two `brock400` instances, and NuMVC$_2$ also fails to solve neither of them.

Although NuMVC$_2$ shows competitive performance with NuMVC, its performance is given by optimizing the $pd$ parameter for each instance. Moreover, as with DLS-MC (Pullan & Hoos, 2006), NuMVC$_2$ is considerably sensitive to the $pd$ parameter. For example, our experiments show that on the `frb53-24` instances, NuMVC$_2$ performs quite well with $pd = 10000$, but it fails to find an optimal solution when $pd$ is set to be a value less than 7000. Comparatively, NuMVC with the same parameter setting performs quite well on all types of instances but the `brock` family. Actually, we will show in the next section that NuMVC is not sensitive to its parameters.

It is also interesting to compare NuMVC with its alternatives which replace the forgetting mechanism with the smoothing techniques similar to those in local search for SAT. Indeed, earlier versions of NuMVC did use the smoothing techniques similar to those in SAT local search, and they did not have good performance compared with NuMVC. It would be interesting to find out the reasons for the success of the forgetting mechanism and the failure of those smoothing techniques in MVC edge weighting local search algorithms such as NuMVC.

## 7.4 Parameters for the Forgetting Mechanism

| | brock400_2 | MANN_a45 | C4000.5 | frb53-24-1 | frb53-24-2 | frb56-25-1 | frb56-25-2 |
|---|---|---|---|---|---|---|---|
| $(0.3\lvert V\rvert, 0.1)$ | 100% (382) | 100% (153) | 100% (262) | 80% (904) | 100% (348) | 100% (338) | 80% (997) |
| $(0.3\lvert V\rvert, 0.2)$ | 100% (361) | 100% (164) | 100% (265) | 85% (918) | 100% (279) | 90% (671) | 70% (1197) |
| $(0.3\lvert V\rvert, 0.3)$ | 100% (362) | 100% (131) | 100% (272) | 70% (1058) | 100% (156) | 95% (826) | 85% (819) |
| $(0.3\lvert V\rvert, 0.4)$ | 95% (490) | 100% (208) | 100% (270) | 80% (995) | 100% (191) | 100% (602) | 100% (885) |
| $(0.3\lvert V\rvert, 0.5)$ | 90% (507) | 100% (90) | 100% (268) | 65% (1316) | 100% (431) | 100% (490) | 95% (922) |
| | | | | | | | |
| $(0.4\lvert V\rvert, 0.1)$ | 100% (261) | 100% (133) | 100% (250) | 80% (899) | 100% (158) | 90% (464) | 70% (1601) |
| $(0.4\lvert V\rvert, 0.2)$ | 90% (736) | 100% (207) | 100% (245) | 80% (860) | 100% (443) | 85% (611) | 80% (851) |
| $(0.4\lvert V\rvert, 0.3)$ | 100% (402) | 100% (176) | 100% (258) | 75% (1047) | 100% (260) | 90% (976) | 90% (1055) |
| $(0.4\lvert V\rvert, 0.4)$ | 95% (375) | 100% (169) | 100% (253) | 70% (1009) | 100% (394) | 90% (885) | 85% (1019) |
| $(0.4\lvert V\rvert, 0.5)$ | 90% (612) | 100% (190) | 100% (264) | 65% (1059) | 100% (137) | 95% (428) | 100% (851) |
| | | | | | | | |
| $(0.5\lvert V\rvert, 0.1)$ | 100% (523) | 100% (107) | 100% (262) | 70% (1007) | 100% (416) | 90% (714) | 75% (1064) |
| $(0.5\lvert V\rvert, 0.2)$ | 85% (950) | 100% (69) | 100% (259) | 75% (1061) | 100% (482) | 95% (706) | 70% (1228) |
| $(0.5\lvert V\rvert, 0.3)$ | 96% (512) | 100% (86) | 100% (252) | 86% (850) | 100% (205) | 100% (470) | 97% (625) |
| $(0.5\lvert V\rvert, 0.4)$ | 90% (499) | 100% (169) | 100% (251) | 70% (931) | 100% (219) | 90% (632) | 80% (1027) |
| $(0.5\lvert V\rvert, 0.5)$ | 90% (968) | 100% (148) | 100% (249) | 90% (805) | 100% (361) | 85% (933) | 85% (983) |
| | | | | | | | |
| $(0.6\lvert V\rvert, 0.1)$ | 100% (527) | 100% (203) | 100% (255) | 70% (1109) | 100% (267) | 100% (828) | 90% (878) |
| $(0.6\lvert V\rvert, 0.2)$ | 80% (713) | 100% (172) | 100% (279) | 75% (944) | 100% (254) | 90% (704) | 70% (1306) |
| $(0.6\lvert V\rvert, 0.3)$ | 75% (976) | 100% (92) | 100% (272) | 70% (1130) | 100% (298) | 90% (689) | 75% (862) |
| $(0.6\lvert V\rvert, 0.4)$ | 100% (710) | 100% (142) | 100% (276) | 75% (907) | 100% (170) | 100% (592) | 85% (1028) |
| $(0.6\lvert V\rvert, 0.5)$ | 85% (742) | 100% (125) | 100% (288) | 80% (947) | 100% (192) | 100% (647) | 80% (1109) |
| | | | | | | | |
| $(0.7\lvert V\rvert, 0.1)$ | 100% (410) | 100% (87) | 100% (273) | 65% (1186) | 100% (358) | 75% (1014) | 75% (934) |
| $(0.7\lvert V\rvert, 0.2)$ | 95% (781) | 100% (128) | 100% (284) | 70% (1035) | 100% (220) | 80% (713) | 90% (510) |
| $(0.7\lvert V\rvert, 0.3)$ | 90% (826) | 100% (125) | 100% (266) | 75% (916) | 100% (206) | 80% (878) | 80% (971) |
| $(0.7\lvert V\rvert, 0.4)$ | 75% (1219) | 100% (101) | 100% (272) | 85% (700) | 100% (338) | 100% (536) | 85% (769) |
| $(0.7\lvert V\rvert, 0.5)$ | 90% (707) | 100% (92) | 100% (280) | 70% (1085) | 100% (352) | 90% (736) | 70% (1044) |

Table 10: Comparative performance of NuMVC with various parameter combinations $(\gamma, \rho)$ for the forgetting mechanism. For each instance, NuMVC is performed 20 times with each parameter combination, except for the one adopted in this work $(0.5\lvert V\rvert, 0.3)$, where the results are based on 100 runs. For `keller6`, NuMVC performs almost the same with various parameters, having the same success rate (100%) and tiny difference of averaged run time (less than 1 second), and thus the results are not reported in the table.

The NuMVC algorithm has two parameters $\gamma$ and $\rho$, which specify the forgetting mechanism. Specifically, when the averaged weight of all edges achieves a threshold $\gamma$, all edge weights are multiplied by a constant factor $\rho$ $(0 < \rho < 1)$. In this subsection, we investigate how NuMVC performs with different settings to these two parameters. The investigation is carried out on both DIMACS and BHOSLIB benchmarks. For the DIMACS benchmark, we select the four instances used in the preceding subsection for the same reasons. For the BHOSLIB benchmark, we select `frb53-24-1`, `frb53-24-2`, `frb56-25-1` and `frb56-25-2`, which are of different sizes and appropriate hardness.

Table 10 presents the performance of NuMVC with various parameter combinations of $\gamma$ and $\rho$ on the representative instances. As we can see from Table 10, the parameter combination $(0.5|V|, 0.3)$ yields relatively good performance for all instances, and exhibits a better robustness over the instances than other parameter combinations do.

On the other hand, we observe that NuMVC with various parameter combinations performs comparably on these tested instances. For example, for all parameter settings, NuMVC achieves a success rate of 100% for `keller6`, `MANN_a45`, `C4000.5` as well as `frb53-24-1`, and the averaged run time difference on these instances is not so significant. For other instances, the difference of success rate never exceeds 25% between any two parameter settings. This observation indicates that NuMVC seems not sensitive to the two parameters. Actually, as we have mentioned before, NuMVC exhibits very good performance for both DIMACS and BHOSLIB benchmarks with a fixed parameter setting. This is an advantage compared to other forgetting mechanisms such as the one used in DLS-MC (Pullan & Hoos, 2006), which is sensitive to its parameter. For algorithms that are sensitive to their parameters, considerable parameter tuning is required in order to get a good performance for a certain instance, which usually costs much more time than solving the instance.

## 8. Conclusions and Future Work

In this paper, we presented two new local search strategies for the minimum vertex cover (MVC) problem, namely two-stage exchange and edge weighting with forgetting. The two-stage exchange strategy yields an efficient two-pass move operator for MVC local search algorithms, which significantly reduces the time complexity per step. The forgetting mechanism enhances the edge weighting scheme by decreasing weights when the averaged weight reaches a threshold, to periodically forget earlier weighting decisions. Based on these two strategies, we designed a slight, yet effective MVC local search algorithm called NuMVC. The NuMVC algorithm was evaluated against the best known heuristic algorithms for MVC (MC, MIS) on standard benchmarks, i.e., the DIMACS and BHOSLIB benchmarks. The experimental results show that NuMVC is largely competitive on the DIMACS benchmark and dramatically outperforms other state-of-the-art heuristic algorithms on all BHOSLIB instances.

Furthermore, we showed that NuMVC is characterized by exponential RTDs, which means it is robust w.r.t. the cutoff parameters and the restart time, and hence has close-to-optimal parallelization speedup. We also performed further investigations to provide further insights into the two new strategies and their effectiveness. Finally, we conducted an experiment to study the performance of NuMVC with different parameter settings, and the results indicate that NuMVC is not sensitive to its parameters.

The two-stage exchange strategy not only has a lower time complexity per step, but also has the flexibility to allow us to employ specific heuristics in different stages. An interesting research direction is thus to apply this idea to other combinatorial problems whose essential tasks are also to seek for an optimal subset with some fixed cardinality.

## Acknowledgments

This work is supported by 973 Program 2010CB328103, ARC Future Fellowship FT0991785, National Natural Science Foundation of China (61073033, 61003056 and 60903054), and Fundamental Research Funds for the Central Universities of China (21612414). We would like to thank the editor and anonymous reviewers for their valuable comments on earlier versions of this paper. We would also like to thank Yanyan Xu for proofreading this paper.